# Should *Terminology Principles* be re-examined?

Christophe Roche

Condillac Research Group - Listic Lab.
Campus Scientifique - Université de Savoie - France
roche@univ-savoie.fr

**Abstract.** Operationalization of terminology for IT applications has revived the Wüsterian approach. The conceptual dimension once more prevails after taking back seat to specialised lexicography. This is demonstrated by the emergence of ontology in terminology. While the *Terminology Principles* as defined in Felber's manual and the ISO standards remain at the core of traditional terminology, their computational implementation raises some issues. In this article, while reiterating their importance, we will be re-examining these *Principles* from a dual perspective: that of logic in the mathematical sense of the term and that of epistemology as in the theory of knowledge. We will thus be clarifying and describing some of them so as to take into account advances in knowledge engineering (ontology) and formal systems (logic). The notion of *ontoterminology*, terminology whose conceptual system is a formal ontology, results from this approach.

**Keywords:** Terminology, Terminology principles, ISO standards, ISO 704, ISO 1087-1, Knowledge theory, Ontology, Logic, Ontoterminology.

## 1 Introduction

Nowadays, terminology, as an independent discipline, risks being absorbed into specialised lexicography or knowledge engineering; the former reducing it to a study of linguistic phenomena and the latter, to an issue of computational knowledge representation. Nevertheless, terminology as a scientific discipline is crucial if we consider that its primary aim is to understand the world, describe the objects that populate it and find the right words to talk about them. Although terminology aims to clarify communication between humans and not to provide computational models [16], we are forced to recognise that it is not entirely satisfactory from the perspective of either Logic (by providing consistent definitions), computation (through a conceptual system representation) or even in respect of epistemological principles (the *essential characteristic* is no longer a principle in the latest version of the ISO 704 standard [2009]). If terminology is to continue to exist as an independent scientific discipline, it needs to re-examine its *Terminology Principles* [29].

This is not intended as yet another criticism of traditional terminology [15], [7], [33]. On the contrary, our contribution is distinctly Wüsterian in its scientific ap-proach. The *Terminology Manual* [10] as well as the ISO 704 standard [16] rightly state that Terminology is multidisciplinary and draws support from a number of disciplines, e.g. logic, epistemology, philosophy of science, linguistics, etc. Nevertheless a lot a work remains to be done in order to really take into account the lessons learnt from these disciplines. Furthermore, some current thinking and practices in the field of terminology must be integrated into a reviewed version of the ISO 704 standard, in particular to take the notion of ontology for terminology into consideration [26], [27], [30]. In examining the basic premises of terminology, we would like to demonstrate the key role of Logic (terminology is a science), epistemology (essential to understanding the world) and, a more recent development, computational models (IT applications require terminology to be operationalized). To this end, we will study the *Terminology Principles* with respect to conceptual system construction in the demanding framework of formal systems[1] – while term definition is dealt with using natural language, concept definition requires a formal language. We will limit our study to the notions of *object*, *concept*, *characteristic* and *relation*. Insofar as possible, we will try to understand why certain issues arise and when appropriate, suggest more accurate definitions for the *Principles*. We will also attempt to align its vocabulary with that of Logic and Knowledge engineering, both of which are compulsory disciplines nowadays.

NB: By *Terminology Principles,* or *Principles* for short, we mean the study of concepts and their relations. In this article, we refer to the Wüster's works [40], to the *Terminology Manual* [10] and to the ISO 704 and 1087-1 Standards on Terminology [16], [18].

## 2 Reality and Object

"Producing a terminology requires an understanding of the conceptualization" [16]. It means on one hand, understanding the "reality" and on the other, organizing the objects that populate it - two different mind operations that are all too often confused. We must bear in mind some epistemological principles (see the note at the end of this paragraph). One of them is that there are two different kinds of knowledge. "An object is defined as anything perceived or conceived" [16]. It defines the first level (and first kind) of knowledge[2] called *individual* (or *singular*) *knowledge* (*object* or *individual* for short): "Such things are called individuals because each thing is composed of a collection of characteristics which can never be the same for another; for the characteristics of Socrates could not be the same for any other particular man" [21] - *individual* must not be confused with *singular concept* which "is said of only one thing". The second kind of knowledge is *conceptual knowledge*, *concept* for short, i.e. *knowledge about a plurality of things verifying the same law*: "which are predicated of many things" [21]. Another fundamental epistemological principal is about the nature of characteristics, some are *essential* when others are *descriptive*. Definition

[1] i.e. systems such as Logic with clearly defined syntax and semantics.
[2] We access an object only through a representation.

relies on the former whereas description relies on the latter. Let us take the ISO 704 example of pointing devices in computer hardware [16]. It is evident that "having a ball on its underside" and "having colour" have not the same importance and do not play the same role in the modelling process.

NB: Most of the "epistemological" quotations are from the "Isagoge" of Porphyry [21], a foundation book for every terminologist and knowledge engineer [31]. The Isagoge epistemology relies on the principle of *difference*: "every difference added to something modifies it" [Isagoge 8.15-20]. Three kinds of differences are required: "Three species of difference have been observed: (1) the separable and (2) the inseparable, and of the latter (2a) the inseparable *per se* and (2b) the inseparable by accident" [Isagoge 9.25] The following figure links the principles of terminology and knowledge engineering with the various kinds of differences.

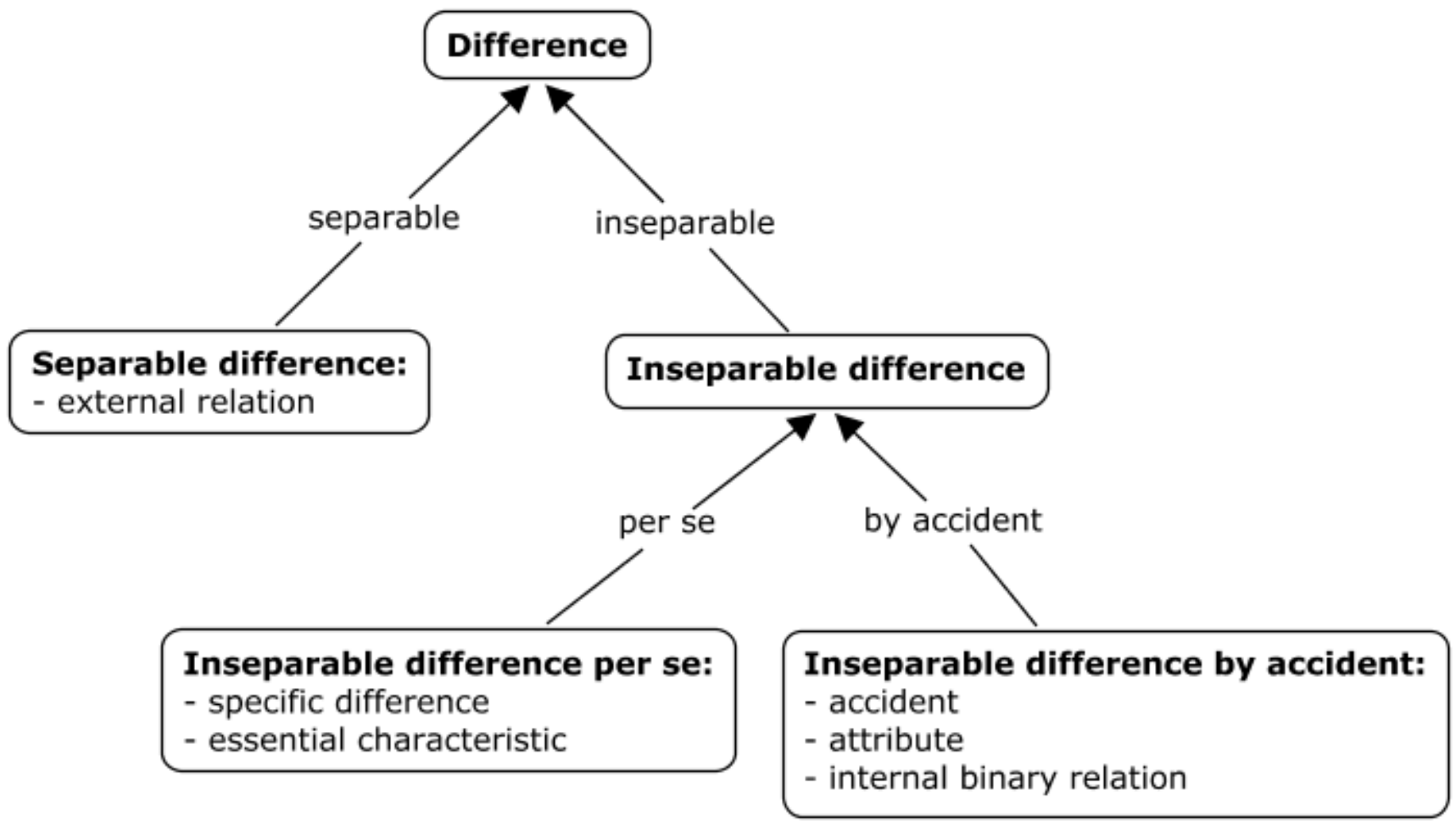

**Fig. 1.** The three kinds of difference

## 3 Concept

"*Objects* are categorized into classes, which correspond to units of knowledge called *concepts*" [16]. Since "concepts are the basis of all terminological work" [10], the first task is to define what we mean by *concept*.

The ISO 704 standard distinguishes two kinds of concepts. The *individual concept* "which corresponds to only one object" [18] and the *general concept* "which corresponds to two or more objects which form a group by reason of common properties" [18]. Such a distinction is not useful and should be avoided. As a matter of fact a concept is a unit of knowledge which involves a plurality of things whatever the number of objects it depicts (one, two or more, or even zero[3]). The definition of a concept does not depend on the cardinality of its extension. Furthermore, considering any unique object as an *individual concept*[4] is not only unnecessary in terminology work but it raises several issues. If "Canada" is an *individual concept*, what are its *delimiting characteristics* from its *generic concept* "Country"? Do we need to create a new individual concept every time the population of Canada changes? An object is a subject for predicates ("To be is to be the value of a variable" [22]), not a predicate. It is a *reification*[5] of a concept. An individual remains itself in essence in spite of its possible variations in quality.

## 4 Characteristic

A concept is a "unit of knowledge created by a unique combination of *characteristics*" [18]. A characteristic is an "abstraction of a property of an object or of a set of objects" [18] i.e. a unary predicate on individuals, e.g. "having colour". Let us notice that the "having colour" characteristic implies that there are as many different characteristics as there are particular colours: "is blue", "is red", etc. All these characteristics are independent when they are semantically linked. We will see that a certain kind of *object properties*, those which describe the state of an object, will be more correctly represented in terms of attributes-values pairs or of binary predicates (binary relations). For example, the "colour" attribute of *this* optical mouse whose value is "blue" corresponds to the following "Colour" binary predicate:
Colour (*thisOpticalMouse*, "blue").

*Characteristics* play an essential role in terminology: "Characteristics shall be used in the analysis of concepts, the modelling of concept systems, and in the formulation of definitions" [16]. The standard adds: "Similarities between *concepts* are indicated by shared *characteristics*; differences that set a concept apart are signalled by *delimiting characteristics*" knowing that "The same *characteristic* of a *concept* may be delimiting in relation to one related *concept* but shared with another related *concept*."

It is evident that all characteristics are not equivalent. In the modelling of the pointing device system, "having a ball on its underside" and "having colour" have not the same importance and do not play the same role in the modelling process. The former participates to the nature of the object – it is an *essential characteristic* – when the latter only describes it. Under these conditions, it is surprising that the latest version of ISO 704 [16] removed the *essential characteristic*: a "*characteristic* which is indispensable to understanding a *concept*" [17]. The reasons put forward are that "An *essential characteristic* is one of a set of *characteristics* that is both necessary and sufficient to determine the *extension* of a concept" and "*Terminology work* is con-cerned with the *intension* and *designation* of *concepts*, and in this context necessary, sufficient, and *essential* characteristics are not used." [16]. But if an essential characteristic is necessary, it is not necessarily sufficient. For example, "mortal" is an *essential characteristic* of a "Human being" since if you remove the "mortal" characteristic from a "Human being", he is no longer a "Human being". Thus "mortal" is necessary, but is not sufficient since "Animal" is also mortal. Furthermore, as it is said in the same document, "the *characteristics* making up the *intension* […] determine the *extension*". So, does it mean that Terminology work does not use *characteristics*? Of course not. The *essential characteristic* as defined in the previous version of ISO 704 and in ISO 1087-1 must be reintroduced in a reviewed version of ISO 704. Finally, does it imply, as it stated in ISO 1087-1, that a *delimiting characteristic* is necessarily an *essential characteristic* ("*essential characteristic* used for distinguishing a *concept* from related concepts")? Not necessarily, as we are going to see.

---

[3] e.g. "the first man on Mars" whose extension is until now empty.

[4] *Singular concept* is involved in theories like individuation, meaning and reference, and definite description.

[5] Called "instance" in knowledge engineering.

Thus, there are two types of characteristics, those which are essential, i.e. "indispensable to understanding a concept" [18], and those which are not. The *definition* relies on the former whereas the *description* relies on the latter.

The characteristics which make the subject *different* are called *essential characteristics*: "Differences present *per se*, then, are comprehended in the definition of the substance and make another essence" [Isagoge 9.15]. They are essential in the sense that if they are removed from the subject the latter would no longer be what it is. As such, they cannot be subjected to "a more or a less"[6]: the essence does not vary. Essential characteristics define concepts – it is the Aristotelian definition – and organise these concepts into a system: "Thus, from essential differences the divisions of genera into species arise and definitions are expressed, since they are composed of a genus and such differences" [Isagoge 9.-5].

The characteristics which do not make the subject different but only change its *description* are called *descriptive characteristics*: "but accidental differences are not comprehended in the definition of the substance and do not make another essence but only a difference in quality" [Isagoge 9.15]. They express *valuated* knowledge[7] which, whether they are present or absent, they do not change the subject's essence even if they give "a more or a less" complete description of it: "These [accidental] differences complete the definition of each thing" [Isagoge 9.20].

Unlike *essential characteristics*, *descriptive characteristics* cannot (should not) be represented as unary predicates[8]. These are *attributes* with which values are associated. Attached to objects, these are *internal binary relations*: "accident is what can be-long or not belong to the same thing […] but always exist in a substratum" [Isagoge 13.5]. More than anything else, the object is a prop for *attributes*.

[6] "Differences *per se* do not permit a more or a less" [Isagoge 9.15-20]

[7] Descriptive knowledge subject to "a more or a less": "accidental differences […] include increase and decrease" [Isagoge 9.15-20], e.g. "being coloured".

[8] A descriptive characteristic puts in relation an individual and a value. It requires a binary predicate (binary relation), e.g. the white colour of my computer mouse will be represented by the following expression: Colour (myMouse, white).

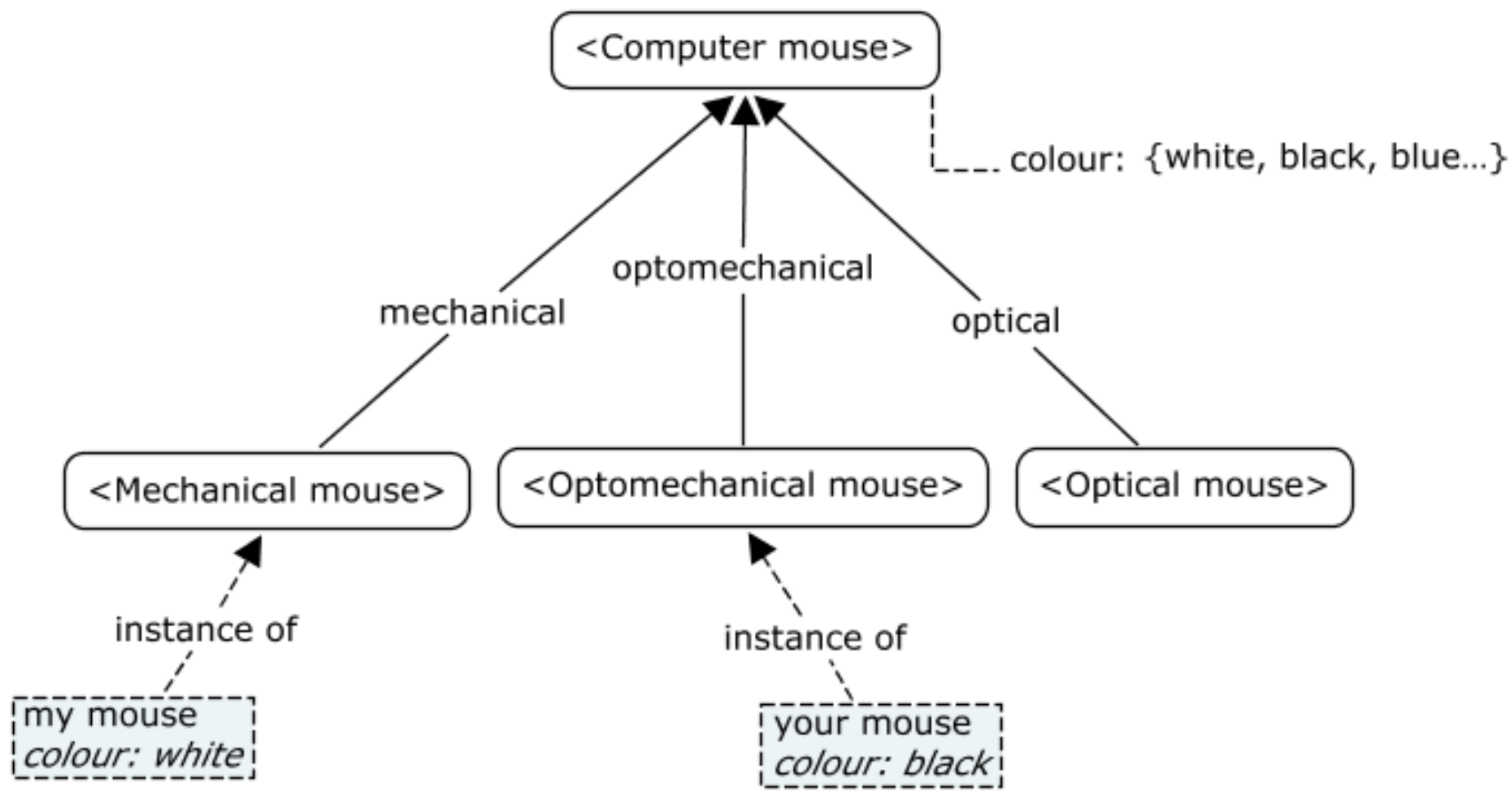

**Fig. 2.** The ISO 704 Computer mouse example

In the field of knowledge engineering, more emphasis is laid on describing objects than on understanding them. The values linked to attributes constitute a number of specific determinations of the object. In modelling stirrers in chemistry, "agitating capacity", "speed range" or "medium's maximum viscosity" are examples of attributes.

This difference between the characteristics is fundamental. As we have just seen, it expresses knowledge of a different nature. *Essential characteristics* define and structure concepts. Those which are not, describe objects and, based on the values linked to them, the various states in which these objects may exist. We shall use the name *attributes* for these *descriptive characteristics* so as to remain in line with the vocabulary of knowledge representation.

ISO standards also introduce the concept of *type of characteristics* to designate characteristics serving as subdivision criteria: "category of *characteristics* which serves as the criterion of subdivision when establishing *concept systems*" [18]. Beyond the issue of barely convincing examples (in many circumstances, "colour" would appear to be more of a quality, i.e. a valuated attribute, than a subdivision criterion), we come up against the issue of managing these characteristics: Are they exclusive? Can they be combined? How do they spread through generic relation? etc. So many questions which require a logical specification of the terminological paradigms.

# 5 Relation

"Concepts do not exist as isolated units of thought but always in relation to each other" [16]. The *Principles* and the ISO standards distinguish between two types of relation between concepts: *hierarchical relations* grouping together *generic* and *partitive relations*, and *associative relations*. The former play a central role insofar as they order the conceptual system and thus enable us to understand and master its complexity: science is the ordering of reality. The latter express a connection considered non hierarchical between concepts "by virtue of experience" [18] such as relations involving a cause and its effect, producer and product, etc.

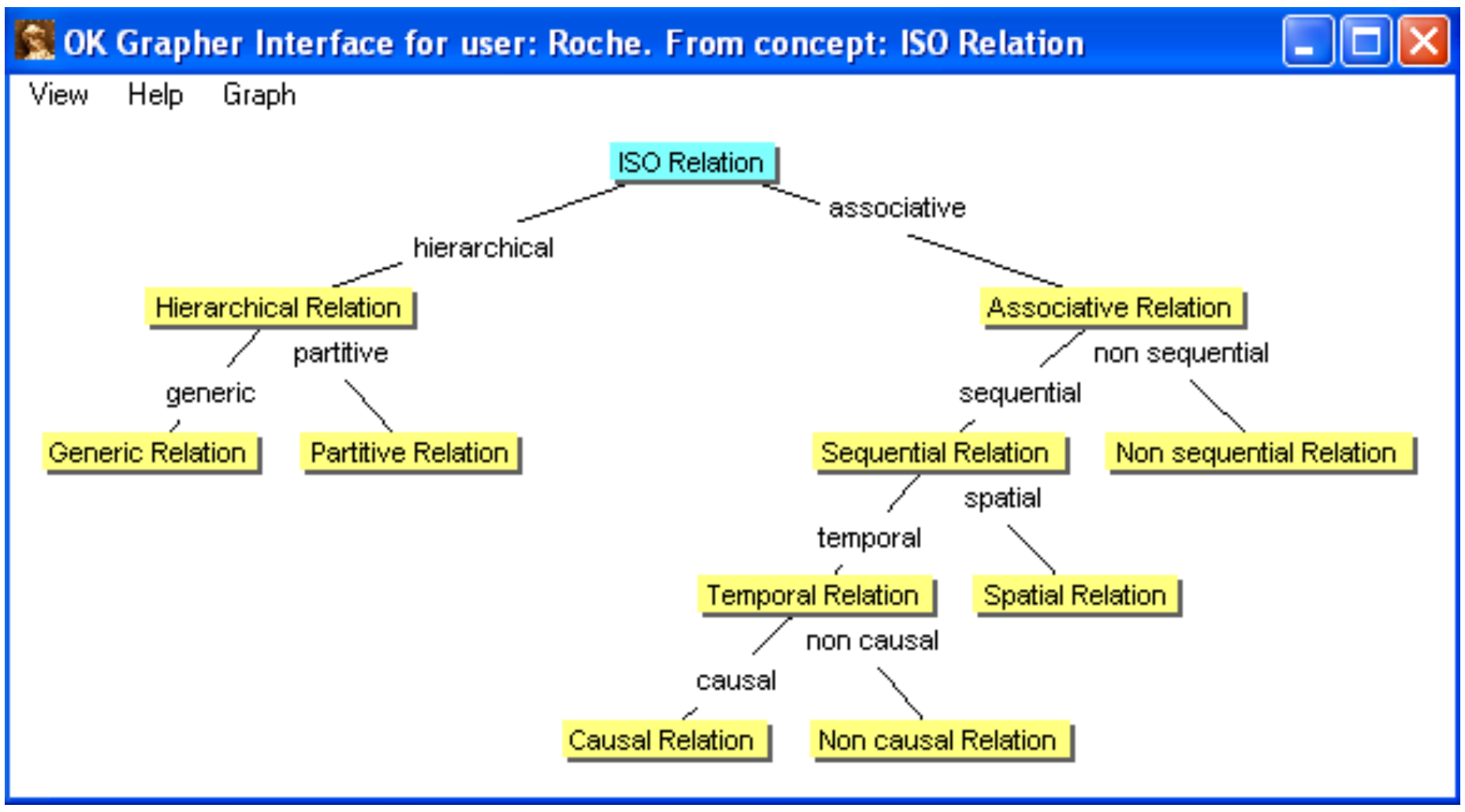

**Fig. 3.** The Porphyry's Tree of Relation[9]

## 5.1 Generic relation

Two concepts are connected by a *generic relation* if the *intension* of the first, named *superordinate concept* and more exactly *generic concept*, is included in the *intension* of the second, named *subordinate concept* and more exactly *specific concept,* and if the latter comprises at least one additional *delimiting characteristic*: "relation between two *concepts* where the *intension* of one of the concepts includes that of the other concept and at least one additional *delimiting characteristic*" [18]. This relation is irreflexive[10], asymmetric[11] and transitive[12]. It therefore defines a *strict order* - a hierarchy - among concepts which is not total insofar as some concepts may not be comparable. *Intension* and *extension* are opposites: the extension of a specific concept is included in the extensions of its generic concepts: “genera have more by containing their subordinate species, while species in their proper differences have more than their genera” [Isagoge 15.20].

---

[9] Built with OCW (Ontology Craft Workbench), an environment for building ontology defined by specific differentiation (© C.Roche, University of Savoie).

[10] Let SCh be the set of characterics, SCp the set of concepts and > the generic relation defined ( $\subseteq$ ) in the Cartesian product SCp×SCp. Saying that C1 is a generic concept of C2, i.e. (C1, C2) $\in$ >, will be expressed by C1>C2. Irreflexive means that a concept can not be subordinated (or superordinated) to itself since a delimiting characteristic is required:

Let us notice that since a concept is defined by a unique combination of characteristics, each element of a partition[13] of this set of characteristics potentially defines a generic concept: the ISO generic relation is poly-hierarchical[14]. Then which superordinate concept immediately above must be used in the intensional definition?

### 5.2 Partitive relation

It is often easier to describe a thing as it is perceived, as it appears to us given its components, than to define it as it *is* (its nature). The *partitive* (part-of) or merological relation therefore plays an important role. It expresses an internal relation between a whole, the *comprehensive concept*, and its components, the *partititive concepts*, without entailing any particular constraints regarding the nature of its constituents. The *Principles* and the ISO standards qualify the *partitive relation* as hierarchical in order to convey the idea that a thing may be understood at increasingly detailed and encompassing levels. However, the *partitive relation* does not define an order[15]. There is no subordination of a component to the whole, in the way that a species is subordinate to a genus. If what is stipulated for the genus is also stipulated for its species, this is not the case for either the whole regarding its components or on the contrary for its components regarding the whole encompassing them.

Considering the partitive relation as a defining and not merely descriptive relation raises numerous issues. This entails the risk of confusing the comprehensive concept's characteristics with its partitive concepts. Even if one can identify “essential” (in the sense of compulsory) and distinctive components (which would differentiate two comprehensive concepts), a partitive concept is not a characteristic – in general, a concept is not a characteristic and by the same token, a characteristic is not a concept[16].

---

$\forall\, C \in SCp,\ \neg\,(C > C)$

[11] $\forall\, C1, C2 \in SCp,\ (C1 > C2) \rightarrow \neg\,(C2 > C1)$

[12] $\forall\, C1, C2, C3 \in SCp, (C1 > C2) \text{ and } (C2 > C3) \rightarrow (C1 > C3)$

[13] A partition of a set S is a set of non-empty subsets of S such that the union of the subsets is equal to S and the intersection of any two subsets is empty. For example, if the concept C is defined by the set of characteristics {c1, c2, c3}, each element of the partition {{c1,c2},{c3}} potentially defines a generic concept

[14] Unlike the generic relation created by the genus–differentia definition

[15] To postulate the transitivity of the part-of relation would vanish the hierarchical levels.

[16] Even if one would like to be able to designate the set of objects having a specific characteristic. This is quite another issue for which Logic provides an elegant solution (see § 7 and 8).

### 5.3 Associative relation

*Associative relations* are *external relations* between concepts (*external* in the sense that they are not necessary to the understanding of the connected concepts) non hierarchical (neither generic nor partitive) by "virtue of experience" [18]. In principle, they do not entail any constraints regarding the nature of connected concepts. *Associative relations* are binary relations, which entails translating relations of greater arity into a set of dyadic relations.

ISO 1087-1 standard gives some examples of associative relations: *sequential*, *temporal* and *causal relations*. The absence of delimiting characteristics in the definition of these relations is regrettable (the figure 3 is a proposal). They could have dissipated certain ambiguities. For instance, is a causal relation also sequential? The given examples suggest this ("action" and "reaction", "nuclear explosion" and "fallout").

### 5.4 Ontological relations

The term "ontology" does not appear in the ISO 704 and 1087-1 standards. The *Terminology manual* [10] speaks of ontological relations to designate indirect relations between concepts and in particular, partitive (merological) relations.

Ontology does not mean the same thing in knowledge engineering which defines it as a formal specification of concepts and their relations[17] [13], [14], [25], [35], [37]. The most important relation is therefore that of subsumption (generic relation) where terminology [10] speaks of a *logical relation* (logical subordination). Given that nowadays, ontology constitutes a theme in itself and is one of the most promising opportunities for modelling and computational representation of the conceptual system of a terminology [26], [30], we suggest using the word ontology in the accepted, knowledge-engineering sense. All the more so given that objects cannot be connected until they have been defined, i.e. until the ontology in the etymological sense of the term has been constructed.

## 6 Definition

According to the *Principles*, definitions are "representations of a concept" [18] and "should reflect the concept system" [16]. Definitions would apparently create[18] nei-ther concepts – these are created by a "unique combination of characteristics[19] " – nor the concept system. Definitions are made afterwards when unique combinations of characteristics have been put together and relations identified.

[17] The first and etymological meaning of ontology is "the science of *being* as *being* independently of its particular determinations". This definition is closer to the one given by knowledge engineering – even if there is some confusion between "beingness" and "existence" – than terminology.

[18] To be compared with the creation of species by its definition in terms of genus and specific difference (the genus-differentia definition).

Here we distinguish two types of definition, intensional and extensional.

### 6.1 Intensional definition

*Intensional definition*, similar to a definition in terms of genus and difference, comprises the *superordinate concept* immediately above followed by one or several *delimiting characteristics*. Yet since the same concept may be directly subordinated to several concepts, which one should be chosen[20]? The fact that different definitions may lead to a same unique combination of characteristics is not a problem in itself. The *Principles* also require specification of the characteristics distinguishing a concept from its coordinate concepts. The advantage of this requirement is even less obvious given that the delimiting characteristics of coordinate concepts with respect to their common superordinate concept automatically distinguishes the coordinate concepts from each other[21] (or at least should do in a well-thought out conceptualisation system).

Definitions based on a partitive or associative relation are not satisfactory from a formal perspective inasmuch as the development of the unique combination of characteristics defining the concept thus created is not clearly determined. What does a delimiting characteristic indicate in the case of a partitive relation? Can the combinations of characteristics of a whole and its components be compared? Of components with each other? Of comprehensive concepts with each other within their structure?

### 6.2 Extensional definition

The *extensional definition* of a *generic* or *comprehensive concept* consists in enumerating all its *subordinate concepts* (*specific* in the case of a *generic concept*, *partitive* for a *comprehensive concept*). This type of definition must not be confused with the "extensional definition" of a set (respectively to a concept) in mathematics which consists in enumerating the objects comprising that set (respectively belonging to the concept). For this reason, we will also refer to "enumerational definition" to avoid any confusion.

While the advantage of this type of definition is understandable, it immediately raises the issue of the unique combination of characteristics identifying the concept thus defined. How can it be created based on the combinations of characteristics of subordinated concepts?

---

[19] It should be specified however that while each unique combination of characteristics creates a concept, this does not necessarily entail a meaning concerning field.

[20] In other words, let C1, C2 and C3 $\in$ SCp with C1={a}, C2={b}, C3={a, b}. C1 and C2 are two generic concepts immediately above C3 and produce two different definitions of C3.

[21] Which is what the Aristotelian definition does with genus and specific difference.

### 6.3 Definition *versus* Description

It is important to bear in mind that partitive and associative relations more describe object than define concept. To close this chapter on definition, let us quote *ces Messieurs de Port-Royal* about their "definition of definition" which illustrates how epistemology and logic, as well as distinguishing essential characteristics from descriptive characteristics, are useful: "There are two kinds of definitions: the more exact one, which retains the name definition, and the other, less exact, which is called a description. The more exact definition explains the nature of a thing by its essential attributes, of which the common one is called the *genus*, and the proper one the *difference*. The less exact definition, called a description, provides some knowledge of a thing in terms of the accidents that are proper to it and determine it enough to give us an idea distinguishing it from other things" [2].

## 7 Logical relation

Extensional (enumerational) definition of a generic concept calls to mind the *Terminology manual*'s concept disjunction [10] – defined in this particular case on the basis of concept extensions. The latter raises identical issues relating to the determination of the characteristics of the generic concept thus created. But it is the same idea: the possibility of building new concepts on the basis of existing concepts, be they generic, specific or comprehensive. Concept conjunction, non-existent in ISO standards, is interesting from this point of view. However, defining the resulting specific concept's intension as the combination of intensions entails certain contradictions. How could we interpret a concept stemming from the conjunction of two coordinate concepts (i.e. stemming from a same generic concept) which would be differentiated by a delimiting characteristic (is it possible to be a thing and its opposite at one and the same time?)?

Nevertheless, Logic should be introduced in Terminology. Concepts, characteristics and attributes are logical predicates (unary for the first two and binary for the last one). Such predicates allow to classify[22] objects into different sets which may be distinct, overlap each other or included. Then it is possible to define *classes*[23] gathering

[22] We distinguish conceptualisation whose aim is to understand the "reality" by defining *concepts* from classification whose goal is structuring objects into *classes*.

[23] If a concept can be interpreted as a set (its extension in the mathematical sense, i.e. the set of the subsumed objects), every set does not correspond to a concept. We introduce the notion of *class* in order to express this distinction: *concept* subsumes individuals of same nature when *class* gathers objects verifying the same property whatever their nature or structure.

individuals verifying the same logical property independently of their nature or structure[24].

## 8 Conclusion and Perspectives

The *Terminology Principles* rightly emphasise the importance of field conceptualisation as the basis of terminology – terminology cannot exist without specialised knowledge. However, creating the concept system is challenging. Especially because knowledge which is often tacit and rarely described in scientific and technical documents, needs to be explained, making experts' participation in and contribution to work on terminology absolutely essential [28]. To help us with this task, the *Principles* propose a certain number of paradigms demonstrating a scientific ambition to order reality – a mathematical structuralism – based on connected concept systems.

However, inaccuracies in these paradigms' definition make terminology operationalization difficult and explain knowledge engineering's leader status in this field – from the computational perspective alone, the *Principles* need to be re-examined. Defining the *Principles* should be done with the same meticulousness as these very *Principles* are intended to apply to ordering reality. Using a formal language with clearly defined syntax and semantics is inevitable not only to eliminate any ambiguities but also because conceptualisation is a scientific activity. Representation languages stemming from artificial intelligence will ultimately lead to terminology operationalization on the basis of a logical specification of the concept system, in the same spirit of formalisation and accuracy but using a different, complementary register.

Logic [3] and artificial intelligence languages [34], [4] are primarily representation systems. Their purpose is not to understand[25] the world but to describe it, formally in the case of the former and for computation in the case of the latter. Their use in the framework of concept system construction should be based on epistemological principles which remain to be specified. These principles should be determined with this objective of formalism and operationalization in mind. To conclude, we shall suggest a few possibilities for achieving this aim.

[24] For example, Red(x) = { x / Couleur (x, red) } gathers all the objects, whatever their nature (concept) and structure (attributes) since they own an attribute "colour" whose value is "red", e.g. my uncle's Ferrari, the apple of my today lunch, etc. Let us notice that the logical property "Couleur (x, red)" is an essential property of the set (its intensional definition) but not of its members.

[25] As with all languages, formal languages divide reality according to their own particular structures. First-order logic achieves this in terms of predicates to which formally defined calculations are applied (predicate calculation). However, such a language cannot directly translate certain fundamental differences. It uses the same formalism to represent both essential (for example Man(x)) and accidental (Sick(x)) characteristics.

Grasping the variety of objects comprising reality is possible on the basis of the notion of concept as a unit of understanding. To this end, its function would be dual: that of understanding what a thing is and that of describing it. The former defines the thing in terms of essential characteristics[26]. The latter describes it as a unique piece of knowledge in the form of valuated attributes[27, 28]. Essential characteristics are born of reason. They participate in the definition of concepts and organise them into a system – a skeleton – on which are hung the attributes describing the objects as they are perceived and whose values express contingent knowledge. *Definition and description are two notions which it is important to differentiate.* They are found in the typology of relations below (fig. 4). Definitions are not limited to subsequently giving expression to a structure built beforehand. In the same operation, they create[29,30] both the concepts and the structure (the conceptual system).

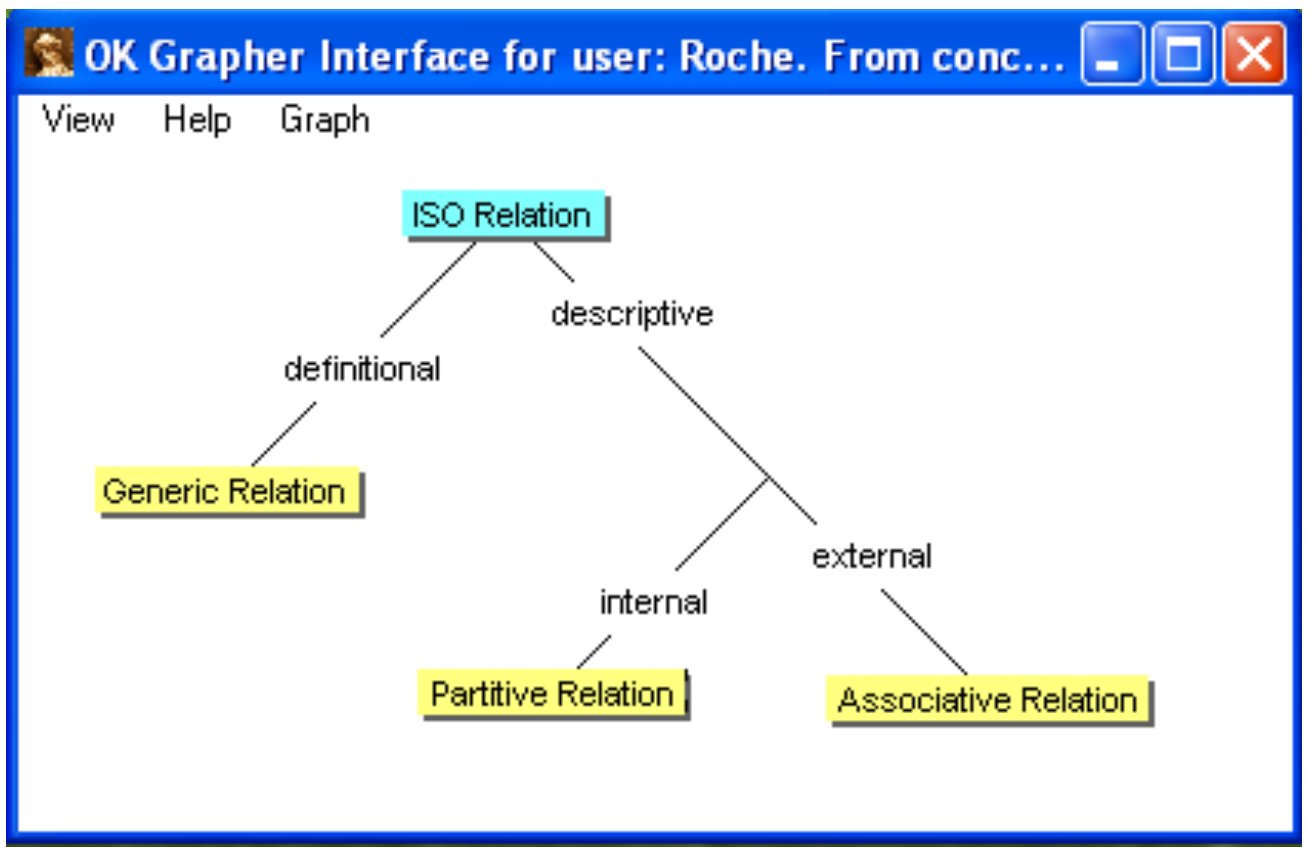

**Fig. 4.** A typology of relations

While every object belongs to a concept[31], it may also be related to different *classes*[32]. Like the *concept*, the *class* is a unit of knowledge concerning a plurality of

[26] A characteristic is considered essential for an object if, on removing it, the object is no longer what it is.

[27] Contrary to an essential characteristic, removing an attribute from an object does not change its nature, it only means it is incompletely described.

[28] We prefer the expression “valuated attribute” to “inherent characteristic” (used in the vocabulary of traditional terminology) so as to emphasise that which is descriptive or contingent in the object.

[29] Provided we consider that a concept’s definition is constructed with respect to previously defined concepts, more knowable than the definiendum.

[30] Unlike a definition in natural language which is rather a linguistic explanation, a formal definition is a *constructive* definition which creates an entity which can be manipulated (in formal and/or computational systems).

[31] to *one* concept corresponding to the object’s essence (from a given point of view).

things. It is however different insofar as the objects it brings together may be of different nature (concept) or of different structure (attribute) as long as they obey the same law. The notion of concept is enhanced by that of logical predicate. It is no longer limited to a unique combination of characteristics but becomes a truth-value function. The concept is predicative [11], as well as characteristic and attribute: Logic is the key. Those predicates can then be combined in an infinite number of ways, through conjunction (allowing a kind of "poly-hierarchy"), disjunction, negation etc. to define a new *class* as a well-formed formula.

The appearance of new paradigms expresses the desire to incorporate all the different sources tapped by terminology. Thus, *ontoterminology* [27], a terminology whose concept system is a formal ontology, emphasises the importance of the epistemological principles governing field conceptualisation – this is the primary definition of ontology. It also emphasises the necessity of a scientific approach to terminology where the expert plays a key role – it involves ontology in its latest definitions where logic and knowledge representation languages are dominant factors. And lastly, it connects the terms (of usage and standardised) to the conceptual model, while distinguishing the term definitions in natural language (linguistic explanations) from concepts' formal definitions (logical specifications).

## References

1. Alexeeva, L.M.: "Interaction between Terminology and Philosophy. Theoretical Foundations of Terminology Comparison between Eastern Europe and Western Countries". Würzburg: Ergon Verlag (2006).
2. Arnauld A., Nicole P.: "Logic or the Art of Thinking". Translated by Jill Vance Buroker. Cambridge University Press (1996)
3. Baader F., Calvanese D., McGuinness D., Nardi D. and Patel-Schneider P.: "The Description Logic Handbook". Cambridge University Press (2003)
4. Brachman R.J., Levesque H.J.: "Readings in Knowledge Representation". Morgan Kaufmann Publishers, Inc. (1985)
5. Budin, G.: "A critical evaluation of the state-of-the-art of Terminology Theory". ITTF Journal, 12. Vienna. TermNet (2001)
6. Cabré T.: "Theories in terminology". Terminology 9:2. pp. 163-199 (2003)
7. Campenhoudt M.: "Que nous reste-t-il d'Eugen Wüster ?". Colloque international Eugen Wüster et la terminologie de l'Ecole de Vienne. Paris, 3-4 février (2006).
8. Costa R.: "Plurality of Theoretical Approaches to Terminology". Modern Approaches to Terminological Theories and Applications. Heribert Picht [ed.]. Serie: Linguistic Insights. Studies in Language and Communication. Vol.36. Berlin-Bern: Peter Lang Verlag (2006)
9. Depecker L., Roche C.: "Entre idée et concept : vers l'ontologie". Revue Langages n°168, décembre 2007, pp. 106-114 (Éditions Larousse) (2007)

---

[32] Classes may be distinct from one another, contain or cover each other, overlap, allowing a less stringent breakdown of reality than concepts. Conceptualisation and classification are two different operations of the mind which are all too often confused.

10. Felber, H.: "Terminology Manual". Unesco (United Nations Educational Scientific and Cultural Organization) – Infoterm (International Information Centre for Terminology) (1984)
11. Frege G.: "Écrits logiques et philosophiques". Éditions du Seuil. Paris.
12. Gomez-Perez, A., Corcho. O., Fernandez-Lopez, M.: "Ontological Engineering: with examples from the areas of Knowledge Management, e-Commerce and the Semantic Web". Asuncion Gomez-Perez, Oscar Corcho, Mariano Fernandez-Lopez, Springer (2004)
13. Gruber, T.: "A Translation Approach to Portable Ontology Specifications". Knowledge Systems Laboratory September 1992. Technical Report KSL 92-71. Revised April 1993. Appeared in Knowledge Acquisition, 5(2):199-220 (1993)
14. Guarino, N., Carrara, M., and Giaretta, P.: "An Ontology of Meta-Level Categories". In J. Doyle, E. Sandewall and P. Torasso (eds.). Principles of Knowledge Representation and Reasoning: Proceedings of the Fourth International Conference (KR94). Morgan Kaufmann, San Mateo, CA: 270-280 (1994)
15. Humbley J.: "La réception de l'œuvre d'Eugen Wüster dans les pays de langue française". Cahier du C.I.E.L. pp 33-51(2004)
16. ISO 704:2009. "Terminology work - Principles and methods". International Organization for Standardization (2009)
17. ISO 704:2000. "Terminology work - Principles and methods". International Organization for Standardization (2000)
18. ISO 1087-1:2000. "Terminology work-Vocabulary-Part 1: Theory and application". International Organization for Standardization (2000)
19. Madsen, Bodil Nistrup & Hanne Erdman Thomsen: "Terminological Principles Used for Ontologies." Managing ontologies and lexical resources. TKE 2008. Copenhagen: ISV. (2008)
20. Pavel, S. & Nolet, D.: "Handbook of Terminology". Minister of Public Works and Government Services Canada. Catalogue No. S53-28/2001(2001).
21. Porphyry. "Isagoge". Translated by Edward W. Warren, The Pontifical Institute of Mediaeval Studies, 1975
22. W. V. Quine.: "On What There Is". Review of Metaphysics 2:21-38 (1948/1953)
23. Rickert H.: "Théorie de la definition". Gallimard
24. Roche C.: "The "specific-difference principle: a methodology for building consensual and coherent ontologies". IC-AI 2001. Las Vegas USA, June 25-28 (2001)
25. Roche, C. "Ontology: a Survey". 8th Symposium on Automated Systems Based on Human Skill and Knowledge. IFAC. September 22-24. Göteborg (2003)
26. Roche C.: "Terminologie et ontologie". Revue Langages, 157, mars 2005, pp. 48-62 (Éditions Larousse) (2005)
27. Roche, C.: "Le terme et le concept : fondements d'une ontoterminologie". TOTh 2007. Terminologie & Ontologie : Theories et applications. pp. 1-22, Annecy. France. 1er juin (2007)
28. Roche, C.: "Saying is not modelling". NLPCS 2007. Natural Language Processing and Cognitive Science. pp. 47 – 56. ICEIS 2007. Funchal, Portugal, June (2007)
29. Roche, C.: "Faut-il revisiter les Principes terminologiques ?". TOTh 2008. Terminology & Ontology : Theories and applications. pp 53-72, Annecy, France, 5 & 6 June (2008)
30. Roche, C., Calberg-Challot, M., Damas, L., Rouard. P.: "Ontoterminology: A new paradigm for terminology". KEOD 2009. International Conference on Knowledge Engineering and Ontology Development, 5-8 October, Madeira (Portugal) (2009)
31. Roche, C.: "Isagoge de Porphyre". Disputatio TOTh. Terminology & Ontology: Theories and applications, pp 23-33, Annecy, France, 26 & 27 May (2011)

32. Sager, J. "A Practical Course in Terminology Processing". John Benjamins Publishing Company (1990)
33. Slodzian M.: "Comment revisiter la doctrine terminologique aujourd'hui". La banque des mots, n°7 (1995)
34. Sowa J.: "Knowledge Representation". Brooks/Cole (2000)
35. Staab, S., Studer, R.: "Handbook on Ontologies". Steffen Staab (Editor), Rudi Studer (Editor). Springer (2004)
36. Temmerman R.: "Towards New Ways of Terminological Description. The Sociocognitive approach". Amsterdam/Philadelphia: John Benjamins (2000)
37. Ushold, M., Gruninger, M.: "Ontologies: Principles, Methods and Applications". Knowledge Engineering Review, Vol. 11, n° 2, June 1996. Also available from AIAI as AIAI-TR-191 (1996)
38. Wright, S.E., Budin, G.: "Handbook of Terminology Management", volume 1 and 2. John Benjamins Publishing Company (1997)
39. Wüster, E.: "The Machine Tool – An interlingual Dictionary of Basic Concepts". London: Technical Press (1968)
40. Wüster E. "Introduccion a la teoria general de la terminologia y a la lexicografia terminologica". Institut Universitari de Linguistica Aplicada (1998)